\documentclass{article}
\usepackage{ijcai22}

\usepackage{times}

\usepackage{soul}
\usepackage{url}
\usepackage[hidelinks]{hyperref}
\usepackage[utf8]{inputenc}
\usepackage[small]{caption}
\usepackage{graphicx}
\usepackage{amsmath}
\usepackage{booktabs}
\title{Learning to Generate Poetic Chinese Landscape Painting with Calligraphy}

\author{
Shaozu Yuan$^1$ \and
Aijun Dai$^1$\and
Zhiling Yan$^1$\and
Ruixue Liu$^1$\and
Meng Chen$^1$ \and \\
Baoyang Chen$^2$\and
Zhijie Qiu$^2$\and
Xiaodong He$^1$ 
\affiliations
$^1$JD AI, Beijing, China \\ $^2$Central Academy of Fine Arts, Beijing, China  \\
\emails
\{yuanshaozu, daiaijun1, yanzhiling1, liuruixue, chenmeng20, xiaodong.he\}@jd.com \\
\{chenbaoyang, qiuzhijie\}@cafa.edu.cn
}

\begin{document}

\maketitle

\begin{abstract}
In this paper, we present a novel system (denoted as Polaca) to generate poetic Chinese landscape painting with calligraphy. Unlike previous single image-to-image painting generation, Polaca takes the classic poetry as input and outputs the artistic landscape painting image with the corresponding calligraphy. It is equipped with three different modules to complete the whole piece of landscape painting artwork: the first one is a text-to-image module to generate landscape painting image, the second one is an image-to-image module to generate stylistic calligraphy image, and the third one is an image fusion module to fuse the two images into a whole piece of aesthetic artwork.
\end{abstract}

\section{Introduction}

Chinese landscape painting, or shan shui (``mountain-water"), is an essential style of traditional Chinese painting which involves or depicts natural landscapes, using a brush and ink rather than conventional paints. Mountains, rivers and waterfalls are common subjects of shan shui paintings. Besides, calligraphy and poetry are usually inseparable from Chinese landscape painting. All three together are regarded as the purest forms of art \cite{murck1991words}. Poetry expresses the thinking and opinion of the artist. Landscape painting constructs the artistic conception. Calligraphy reflects the emotion and inner psychology of the artist. They complement each other and elevate the artistry of artwork significantly.

With the success of deep learning, Generative Adversarial Networks (GANs) \cite{goodfellow2014generative} have been widely applied in artwork generation and made remarkable progress \cite{xu2018attngan,tomei2019art2real,ding2021cogview}. However,  to the best of our knowledge, previous works only focus on a single form of artwork, such as poetry generation \cite{wang2016chinese,guo2019automated,shen2020compose}, calligraphy generation \cite{zhang2018separating,gao2020gan,liu2020maliang}, or Chinese landscape painting generation \cite{lin2018transform,he2018chipgan,zhou2019interactive}, and no one explores the composition of poetic Chinese landscape painting with calligraphy as a whole piece of artwork. Meanwhile, most of the previous approaches \cite{lin2018transform,he2018chipgan,zhou2019interactive} treat the Chinese landscape painting generation as a style transfer problem based on image-to-image translation, which heavily relies on the conditional inputs (e.g. photograph or sketches), thus it is restricted in the number of generated images. Since each of its generation is built upon a single, human-fed input, the traditional style transfer methods tend to produce derivative artworks that are stylistic copies of conditional inputs, which lacks creativity and imagination for the artwork creation.
\begin{figure}[t]
    \centering
    \includegraphics[width=1\linewidth]{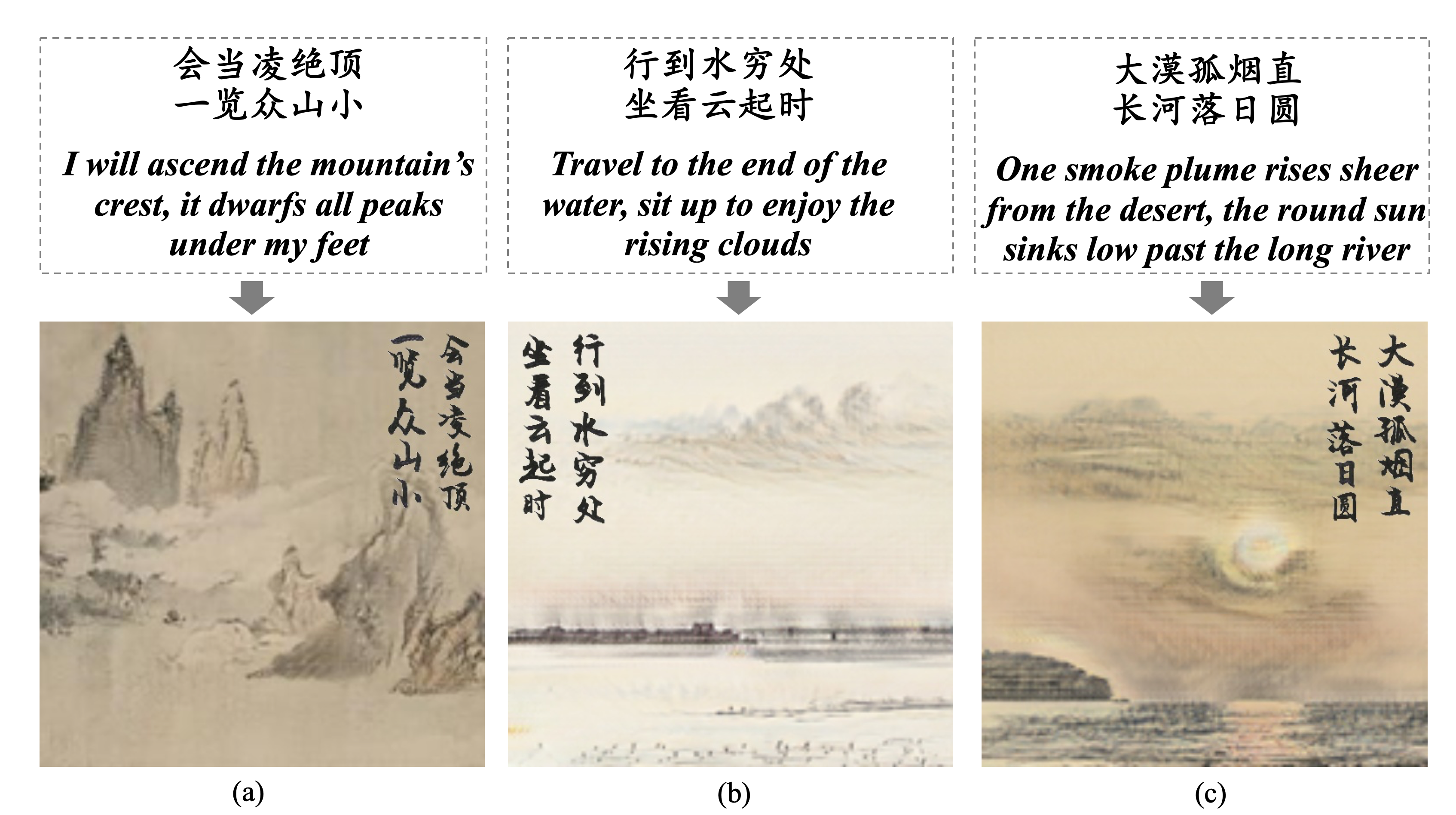}
    \caption{The generated artwork examples of our system.}
    \label{fig:poem imgs}
\end{figure}

\begin{figure*}
\centering
\includegraphics[width=0.95\linewidth]{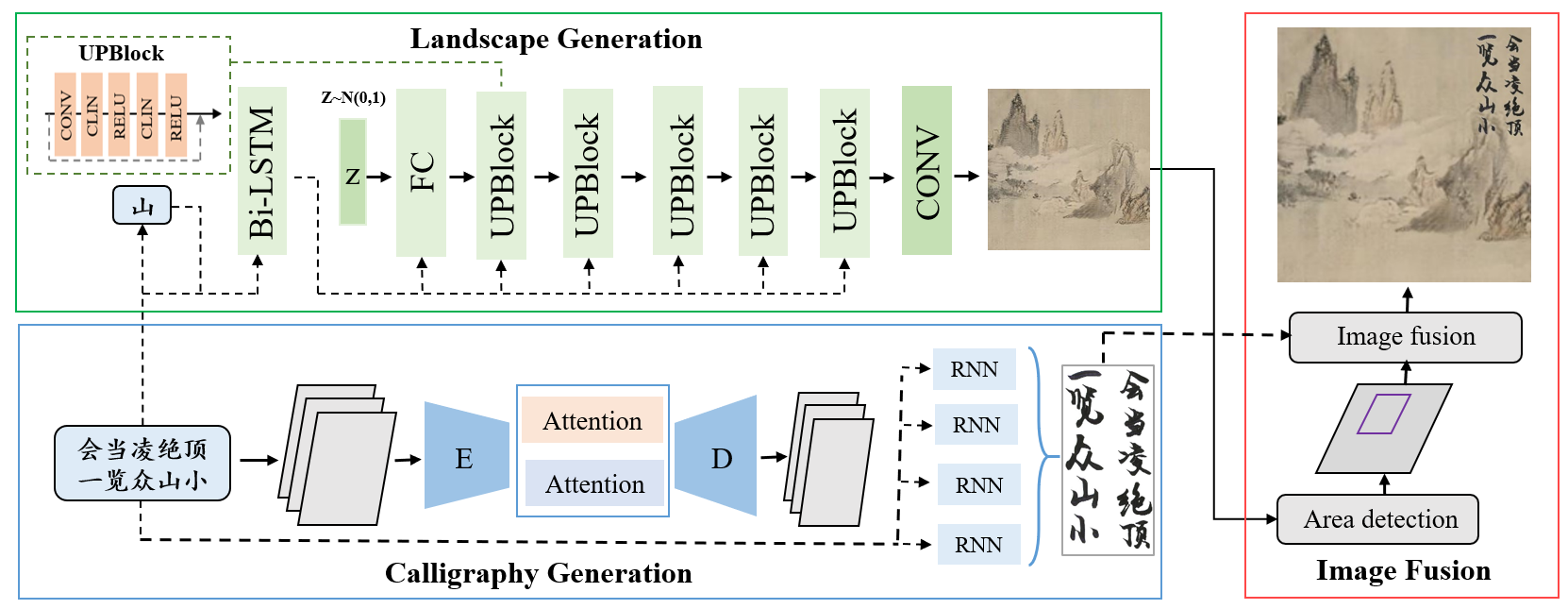}
   \caption{The architecture of our system is composed of three parts: landscape generation, calligraphy generation and image fusion. The input to the system is a classic poetry text, and the output is a landscape painting with calligraphy.}
\label{fig:framework}
\end{figure*}
In this paper, we propose a novel and challenging task of generating poetic Chinese landscape painting with calligraphy\footnote{Demo video: \url{https://youtu.be/xRo8xiTXb74}}. Unlike previous image-to-image approaches, we formulate the task as a text-to-image task and develop a system that takes the classic poetry text as its input and outputs the artistic landscape painting image with calligraphy. Specifically, the generation process is as follows: 
1) Landscape painting generation takes the poetry text as input and applies the text-to-image model to generate a poetic painting image that contains the main subjects of the poetry. 2) Calligraphy generation leverages the style transfer model to generate stylistic calligraphy images from standard font images. 3) The image fusion module predicts the layout of calligraphy and fuses the calligraphy image into the landscape painting image. As there is no off-the-shelf poetry-to-painting dataset, we also construct a dataset with more than 5,000 text-image pairs via an automatic method which will be further introduced later.

\section{Dataset Construction}

We collect some Chinese landscape painting images from previous works \cite{bao2010dataset,xue2021end}. To enrich the data diversity, we also crawl thousands of  real landscape photographs from the web. To construct semantic-related poetry-painting pairs, we set up a dataset construction pipeline as Figure 3 shows. We first train a CycleGAN model \cite{zhu2017unpaired} to transfer the landscape photographs to photographs paintings. To match the classic poetry for each image automatically, we first detect the objects like mountains, rivers and trees in the photograph, then extract key entities (e.g. mountain, water, forest) from poetry via TextRank \cite{mihalcea2004textrank}, finally match the photograph with poetry based on the overlap of entities and objects. Considering the literal differences between ancient Chinese and the output labels of object detection model, we also construct a mapping dictionary based on synonyms to facilitate the matching. By this way, more than 5,000 text-image pairs are constructed for text-to-image model training.


For the calligraphy generation task, we use the same dataset as \cite{liu2020maliang}. We also annotate the location information of calligraphy in 500 traditional landscape paintings to facilitate the training of our image fusion model.   


\begin{figure}[t]
    \centering
    \includegraphics[width=1\linewidth]{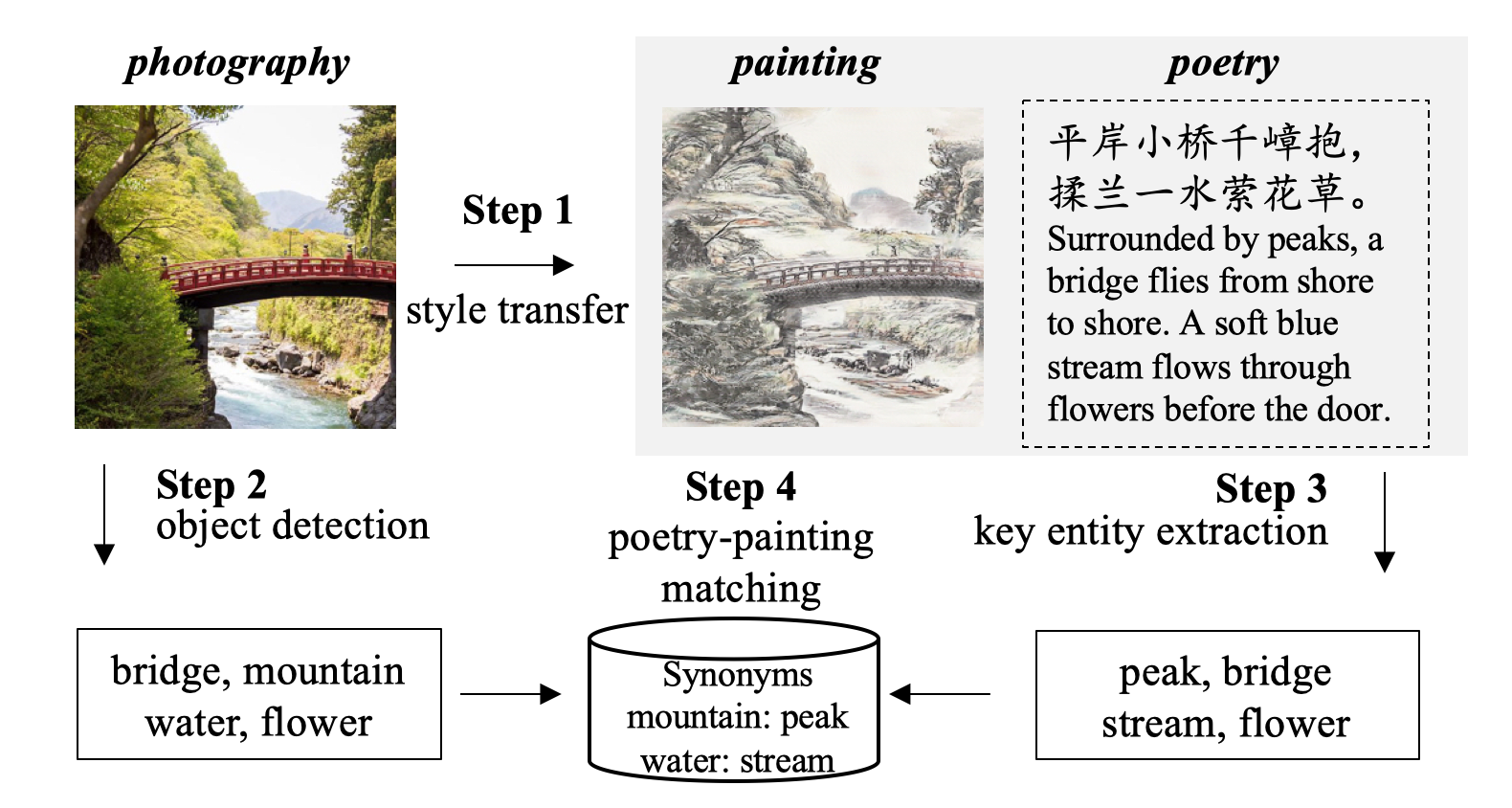}
    \caption{Illustration of dataset construction.}
    \label{fig:dataset}
\end{figure}

\section{System Architecture}
As shown in Figure 2, our system Polaca contains three modules: 
1) a Chinese landscape painting generation module that creates painting images from the input text, 2) a calligraphy generation module that generates calligraphy images based on the input text, 3) a multi-modality image fusion module which combines the generated calligraphy and landscape painting images with area detection and image fusion.

\subsection{Landscape Painting Generation}
To guide the landscape painting generation effectively, we design a novel cross-modal generative adversarial network that is composed of a generator, a discriminator, and a text encoder. The text encoder first encodes the poetry and the extracted basic painting elements into a text vector with Bi-LSTM \cite{huang2015bidirectional}. Then, to ensure the diversity of generated images, we feed the initial image vector, sampled from the Gaussian distribution, into the generator. The text and image features are fused by stacked up blocks during the image generation process. In particular, we adopt CLIN (Conditional Layer-Instance Normalization) \cite{yuan2021learning} in up blocks, which combines the advantages of instance normalization and layer normalization to selectively change or keep the content information. Finally, a discriminator is applied to distinguish real images from synthetic images \cite{goodfellow2014generative}.



\subsection{Calligraphy Generation} 
Calligraphy generation consists of two sub-modules: character image generation and layout prediction. The character image generation module generates a stylistic calligraphy image from standard font images for every character in the input text. To guarantee the quality of the generated character image, we implement a modified GAN model equipped with two auxiliary classifiers with the attention mechanism in the generator to catch both the content-aware representation and style-aware representation by following \cite{kim2019u}. The layout prediction module predicts the overall spatial arrangement for the whole piece of calligraphy artwork. We devise a recurrent neural network \cite{GRU} to solve the sequential modelling problem.

\subsection{Image Fusion Module} 
To fuse the calligraphy image into the landscape image naturally, it is necessary to detect the proper area in the landscape painting image.  Here, we devise a layout network based on Faster R-CNN \cite{ren2015faster} to predict the location and size of the calligraphy image. Then we synthesize the final artwork by implementing pixel-level image fusion. With the advantage of Faster R-CNN and pixel operation, the image fusion module can insert the generated calligraphy image into the generated landscape painting appropriately.   

\section{Evaluation and Analysis}

\subsection{Human Evaluation}
We conduct human evaluation to measure the quality of generated landscape paintings with calligraphy. We first sample 100 real landscape painting images created by human from our dataset, then mix it with 100 generated paintings together to form the test set. Then we ask the human evaluators to score each painting by hiding the image source. Twenty students majoring in fine arts are guided to evaluate the images based on the criteria of landscape quality (landscape), calligraphy quality (calligraphy), coherence between landscape images and poetry (coherence), and aesthetics of the overall artwork (aesthetics). The rating score ranges from 1 to 10 where 10 represents the best quality. Table \ref{table:user} demonstrates that even though the generated images are rated lower than real artworks in the quality of \textit{landscape} and \textit{calligraphy}, they obtain comparative scores with human paintings in the aspect of \textit{coherence}, indicating our proposed architecture can generate semantic-related landscape paintings by understanding the poetry content. 

\begin{table}[h]
\small
\centering
\setlength{\tabcolsep}{1.8mm}{
\begin{tabular}{l|ccccc}
  \hline
  Scores & Landscape & Calligraphy & Coherence & Aesthetics \\ \hline
  Generated & 8.55 & 8.83 & 7.12 & 8.08  \\
  Human & 9.23 & 9.15 &7.22 & 9.12 \\
  \hline
\end{tabular}}
\caption{The user study on generated and real paintings.}
\label{table:user}
\end{table}
\subsection{Case Study}
\begin{figure}[t]
    \centering    \includegraphics[width=1\linewidth]{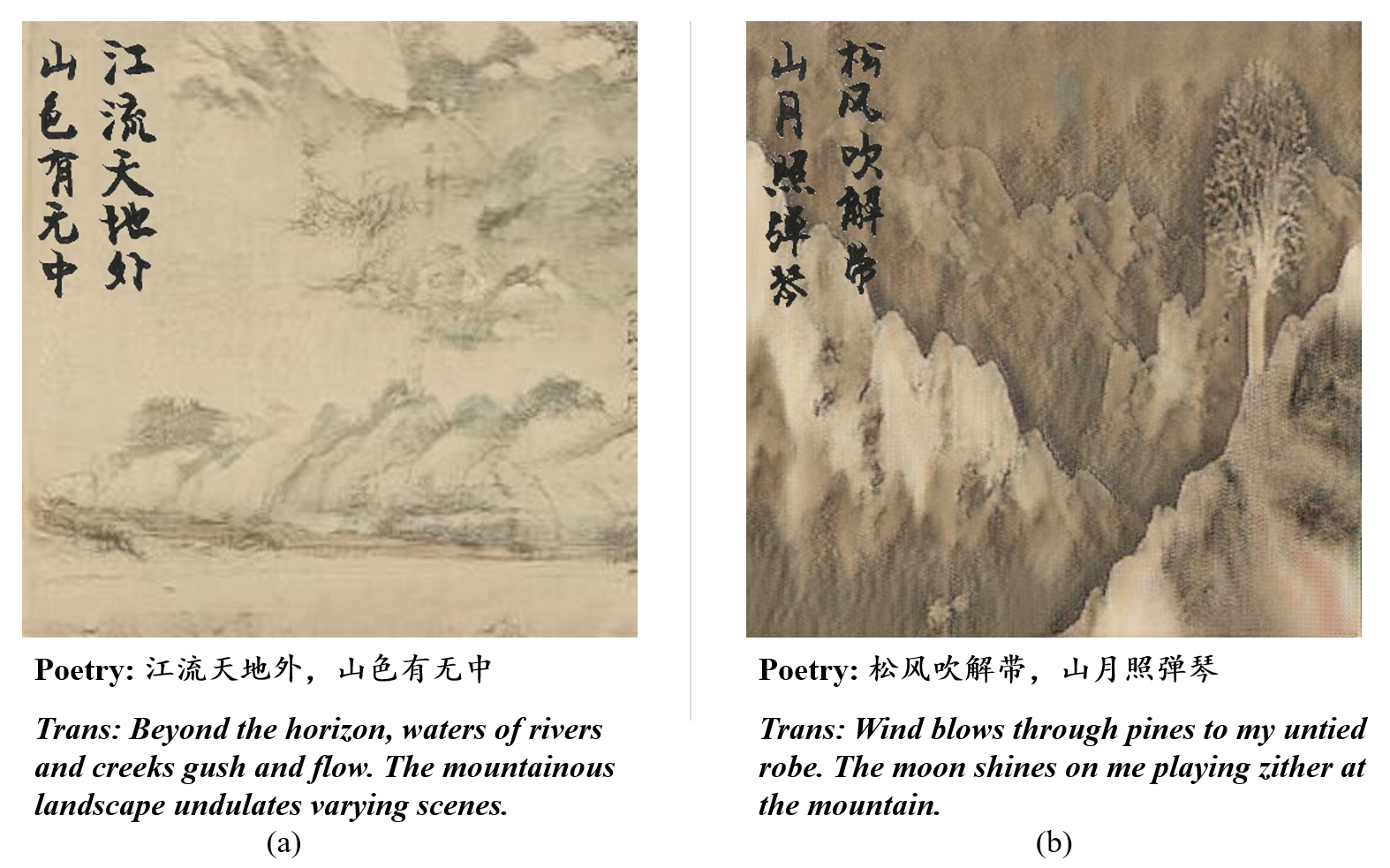}
    \caption{Case study examples.}
    \label{fig:caseStudy}
\end{figure}

Besides, we also make further case study to discuss the advantage and limitation of our system. As shown in Figure \ref{fig:caseStudy} (a), our system captured the key elements of \textit{river, creek, mountain} and depicted them in the generated paintings accurately. And the calligraphy is also visually pleasing. We conjecture that all three elements appear frequently in traditional Chinese paintings and poetry which are easy to learn. Meanwhile, it indicates the effectiveness of the CLIN block on catching the interactions of text and image. However, in Figure \ref{fig:caseStudy} (b), our model omitted the \textit{moon} and \textit{zither} by mistake. What's more, the hues are also too dark, which damaged the artistic conception of whole painting. In the future, we plan to design more objectives to encourage the model to improve the completeness during text-to-image generation.

\section{Demonstration}
\begin{figure}[t]
 	\centering 
 	\includegraphics[width=0.9\linewidth]{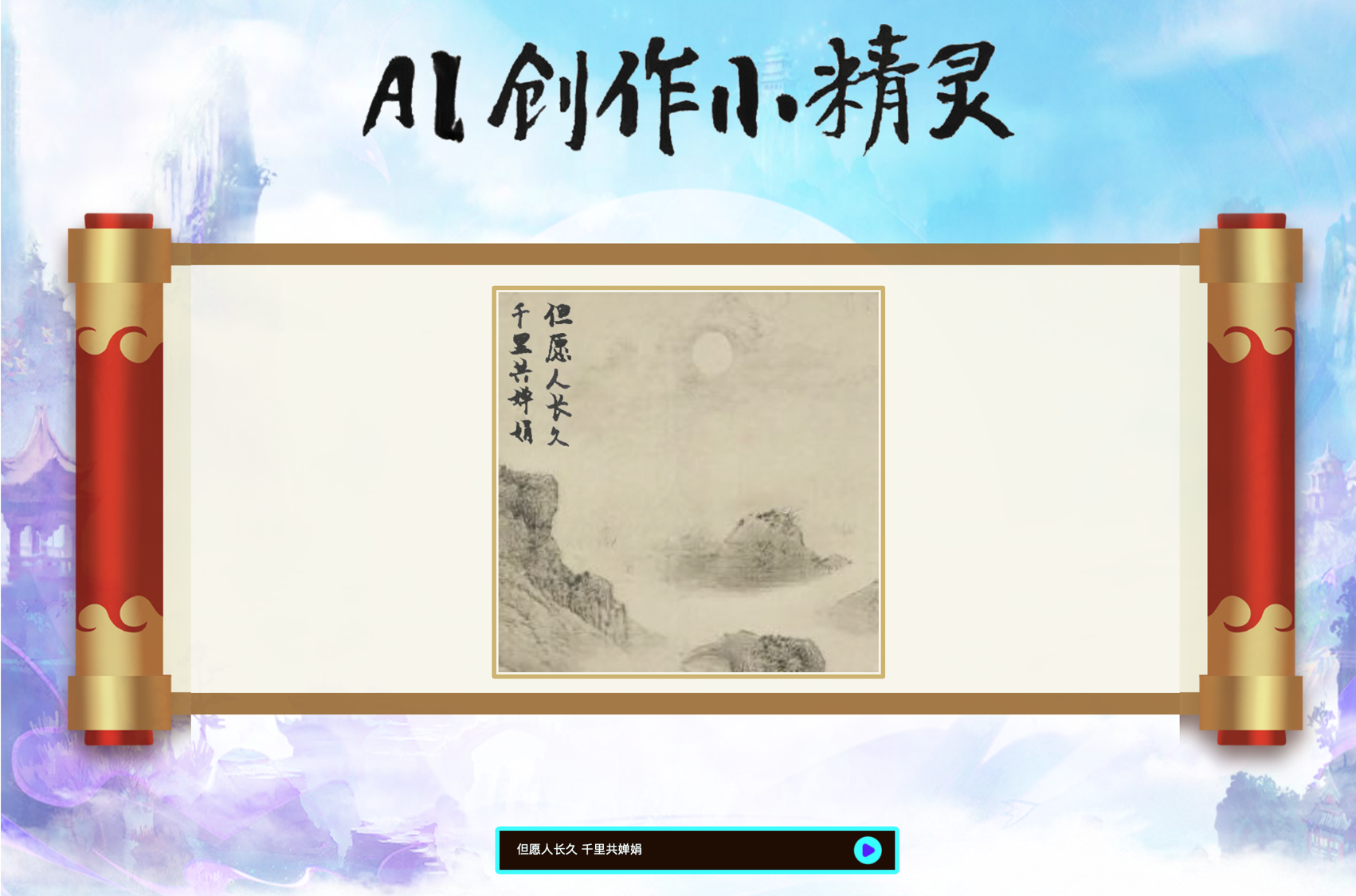}
 	\caption{The interface of our Demo.}  
 	\label{fig:demo}   
 \end{figure}

In order to facilitate a user-friendly experience, we develop a website for users to experience the journey of art creation. Users can create poetic Chinese landscape paintings with calligraphy by entering Chinese poetry. The system is implemented with TensorFlow and Python, and 4 GPUs (NVIDIA Tesla P40) are deployed for real-time inference. Please watch the demonstration video for more details. Figure \ref{fig:demo} shows the demo interface.
 
\section{Conclusion}
In this paper, we develop a novel system to challenge the task of composing poetic Chinese landscape painting with calligraphy. The system consists of a text-to-image module to generate landscape painting, an image-to-image module to create stylistic calligraphy images, and an image fusion model to combine the calligraphy and landscape painting images together naturally. We also contribute a large-scale poetry-to-painting multi-modal dataset. A web-based demo is established to make the system easily accessible. 

\bibliographystyle{named}
\bibliography{ijcai22}

\end{document}